\documentclass[aps,prmaterials,reprint,groupedaddress,longbibliography,showkeys]{revtex4-2}

\usepackage[version=3]{mhchem}
\usepackage{graphicx}
\usepackage[T1]{fontenc}
\usepackage{booktabs}
\usepackage{amsmath,amssymb,amsfonts}
\usepackage{placeins}
\usepackage{hyperref}
\usepackage{xr-hyper}

\begin{document}

\title{A literature-guided descriptor-based framework for filtering composition search spaces}
\thanks{Supplementary Information accompanies this manuscript as a separate PDF.}

\author{Lei Zhang}
\email{lei.zhang-w2i@rub.de}
\author{Markus Stricker}
\email{markus.stricker@rub.de}
\affiliation{Interdisciplinary Centre for Advanced Materials Simulation, Ruhr University Bochum, Universit\"atsstra\ss e 150, 44780 Bochum, North Rhine-Westphalia, Germany}

\begin{abstract}
Scientific literature contains latent knowledge about materials behavior, but much of this knowledge is expressed through words, contexts, and recurring associations rather than explicit design principles. This raises a central question: how can large-scale scientific corpora be used for practical problems in materials discovery? Here, we present a literature-guided descriptor-based filtering framework for reducing composition search spaces. For a given performance metric, the framework selects two descriptors from a filtered vocabulary in a literature-trained word embedding model and uses the selected descriptors to construct a Pareto-based filter for candidate compositions. Across the evaluated performance metrics and composition search spaces, the framework filters out an average of 74.27\% of the candidate compositions, with an average best-value error of 1.93\% relative to experimental measurements. Compared with expert-chosen and random descriptors, our performance metric-dependent descriptors provide a more controlled balance between retained fraction and best-value error. These results show that literature-derived embeddings can support intuitive and reproducible filters for narrowing candidate composition spaces while preserving high-performing compositions.
\end{abstract}


\maketitle

\section{Introduction}

The search for (new) useful materials often begins with a composition search space that is too large to screen brute-force~\cite{butler2018machine,curtarolo2012aflow}. This challenge is especially important for multicomponent systems, where even a modest set of elements can produce a large number of possible candidate compositions, most of which have not been synthesized, reported, or characterized experimentally. In electrocatalyst screening, the same challenge appears in a practical form~\cite{Ludwig2019}. A composition-spread materials library provides an experimental realization of a selected region of composition space, but only a limited number of candidate compositions can be examined in detail by experiment or by more expensive computation~\cite{jain2013materialsproject,chen2025accelerating,zakutayev2017highthroughput}. The central task is therefore not only to locate the best-performing composition for a chosen performance metric, but to narrow the original composition search space to a smaller, experimentally tractable subset while retaining high-performing candidate compositions.

The scientific literature provides one source of guidance for composition-space filtering. Papers do not only report quantitative measurements of performance metrics such as activity, resistance, and stability. Scientific texts also accumulate descriptive vocabulary that connects composition, performance, structure, stability, and function. As a whole, the scientific corpus encodes latent knowledge about which tokens appear in similar contexts and which descriptions tend to be associated with particular materials behavior. Word embedding approaches quantitatively represent these contextual correlations by mapping terms, or tokens, into a continuous vector space derived from co-occurrence patterns in text~\cite{mikolov2013phrases}. Earlier studies, including our own work, have shown that such embeddings can recover nontrivial chemical and materials relationships from the scientific literature alone~\cite{tshitoyan2019materials,qu2024language,jiang2025nlp,Zhang2024,zhang2025electrocatalystdiscoverytextmining,zhang2025compositionproperty}. Literature-derived embeddings can therefore provide more than a retrospective record of prior knowledge; the embeddings can also define directions that help guide the reduction of composition search spaces.

The methodological challenge is how to turn latent knowledge from the scientific literature into a usable and reproducible filter. Previous literature-guided approaches have connected literature embeddings with composition screening in different ways. Word-embedding-derived composition representations have been used as features for surrogate models trained on experimental data~\cite{zhang2025compositionproperty}. In our previous work, expert-chosen terms were instead used as descriptors in embedding space, with candidate compositions represented according to their similarities to these descriptors~\cite{zhang2025electrocatalystdiscoverytextmining}. These studies demonstrate that literature-derived embeddings can support performance-oriented composition screening. However, in the descriptor-based approach, the descriptor choice remains outside the method and may therefore miss useful latent information contained in the embedding space. The selection process also remains empirical and heuristic, i.e., based on trial and error. A workflow based on fixed, expert-chosen descriptors is therefore only partly automatic and may be suboptimal, because the descriptors that define the filter are selected before the performance-specific screening begins.

The limitation of fixed descriptors becomes more important when the performance metric changes. Descriptors chosen by expert knowledge for activity screening may not remain appropriate for resistance or stability screening. These observations suggest that the descriptors should be selected in a performance metric-dependent way rather than imposed in advance. At the same time, the filter should remain interpretable: after selection, the descriptors should be inspectable, comparable across performance metrics, and traceable to the literature-derived embedding from which the descriptors were obtained.

Here we present a literature-guided descriptor-based filtering framework for reducing composition search spaces. For a given performance metric, the framework selects two descriptors from a word embedding space. Two descriptors provide the smallest representation that permits a trade-off-based Pareto filter while keeping the resulting projection directly interpretable. Each candidate composition is represented by its similarity to the first descriptor and its similarity to the second descriptor, and non-dominated candidates are retained in the resulting two-dimensional similarity space. Across ten composition search spaces distributed over four performance metric filtering tasks, our framework reduces the candidate composition space while preserving the best or near-best measured compositions. The main contribution is a reproducible route from latent literature knowledge to an interpretable, performance metric-dependent filter for composition-space reduction in the context of electrocatalysis.

\section{Results}

\subsection{Selecting descriptors in embedding space}
\begin{figure*}[htbp]
\centering
\includegraphics[width=\textwidth]{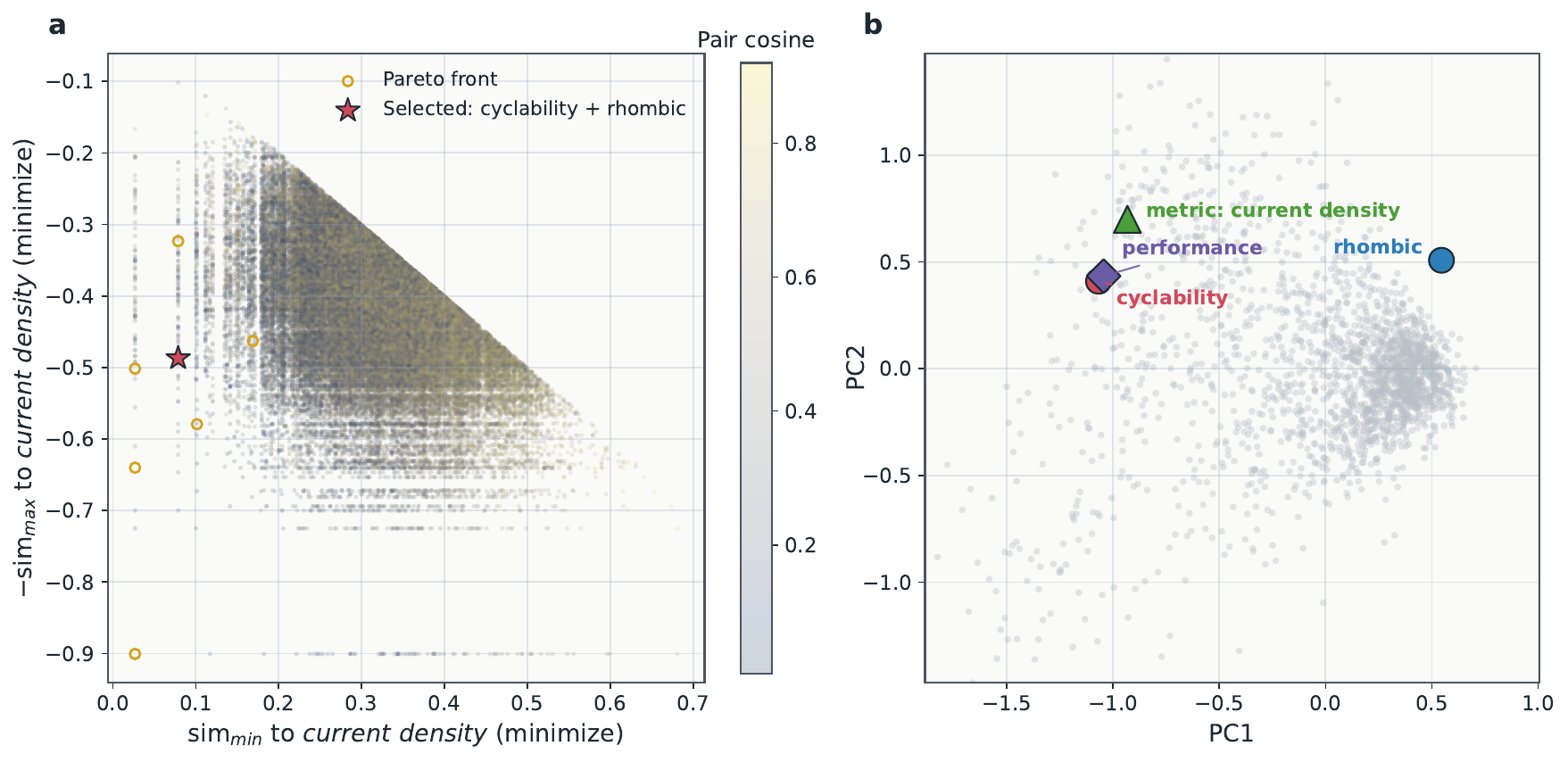}
\caption{Descriptor selection for the performance metric \textit{current density}. (a) Possible two-descriptor combinations in the three-objective selection space. The Pareto-based selection minimizes the lower similarity to \textit{current density}, maximizes the higher similarity to \textit{current density}, and minimizes the cosine similarity between the two descriptors. Open markers indicate the descriptor-level Pareto front, and the selected descriptors are \textit{cyclability} and \textit{rhombic}. Similarity to the word \textit{performance} is not an axis in (a); it is used only in the second stage to select the final descriptor combination from the Pareto front. (b) Two-dimensional PCA projection of the 200-dimensional filtered embedding space, showing \textit{current density}, the word \textit{performance}, and the two selected descriptors.}
\label{fig:concept}
\end{figure*}

All descriptor selection and similarity calculations are performed in the 200-dimensional word embedding space; Figure~\ref{fig:concept}b provides a two-dimensional PCA projection only for visualization.
Before candidate compositions can be filtered, our framework first turns a chosen performance metric into a descriptor-selection problem. The selection is carried out within a candidate descriptor vocabulary derived from the literature-trained word embedding model~\cite{Zhang2024}. Each candidate descriptor is a vocabulary token, and its 200-dimensional word embedding vector is the numerical representation used for selection. The candidate descriptor vocabulary consists of lowercase alphabetic words ending in \textit{-ivity}, \textit{-able}, \textit{-ive}, \textit{-ous}, \textit{-ic}, \textit{-ity}, or \textit{-ce}, as detailed in the Methods (Section~\ref{sec:methods}). The aim of this step is to identify two descriptors per performance metric that can later be used to represent each candidate composition by two corresponding similarity values. For the performance metric \textit{current density}, this procedure identifies \textit{cyclability} and \textit{rhombic} as the most suitable pair (Figure~\ref{fig:concept}).

Descriptor selection proceeds in two steps. First, possible two-descriptor combinations are evaluated by balancing three quantities: the lower and higher descriptor similarities to \textit{current density}, and the cosine similarity between the two descriptors. This design tests the hypothesis that useful filtering requires one descriptor associated with the performance metric and a second descriptor that provides a contrasting, nonredundant view. Accordingly, maximizing the higher similarity keeps one descriptor close to the metric, whereas minimizing the lower similarity and the similarity between the descriptors promotes contrast. Second, because the Pareto front can still contain many possible combinations, the final descriptor combination is chosen by its average similarity to the word \textit{performance}. This second step favors descriptors that remain close to general performance-related language in the literature-derived embedding space.

Figure~\ref{fig:concept}a shows that the possible descriptor combinations cover a broad region of this three-objective selection space. The selected descriptors, \textit{cyclability} and \textit{rhombic}, lie on the descriptor-level Pareto front rather than at an extreme determined by only one objective. The selected descriptors therefore represent a compromise between a close association with \textit{current density}, contrast in descriptor-metric similarity, low similarity between the two descriptors, and average association with the word \textit{performance}.

The embedding-space relation between the selected descriptors and \textit{current density} is shown in Figure~\ref{fig:concept}b. The embedding representation of \textit{current density}, the word \textit{performance}, and the two selected descriptors occupy the same local region of the filtered embedding space. Within that region, \textit{cyclability} lies closer to \textit{current density} and \textit{performance}, whereas \textit{rhombic} is more separated from \textit{cyclability}. The selected descriptors therefore combine one descriptor that is closely associated with the performance metric and a second descriptor that provides contrast.

\subsection{Composition filtering for current density}

After descriptor selection, the selected descriptors are used to filter candidate compositions. For the current density metric, the selected descriptors are \textit{cyclability} and \textit{rhombic} across all five evaluated composition search spaces. The summary across all evaluated performance metrics is shown in Table~\ref{tab:task-summary}, and the detailed composition-filtering results for current density are listed in Table~\ref{tab:si-current-density-prediction}. 

For the current density metric, our framework retains on average $6.87 \pm 2.36$\% of candidate compositions across the five composition search spaces, with an average best-value error of $3.85 \pm 6.38$\%. Best-value error is calculated as the relative difference between the best current density value in the retained subset and the best current density value in the full composition search space; for current density, lower negative values indicate better performance. In three of the five composition search spaces, the retained subset includes the best-performing composition from the full search space. This combination of strong reduction and low best-value error is visible across the individual composition search spaces: Ag-Au-Pd-Pt-Rh is reduced from 327 to 24 compositions while retaining the best-performing composition; Ag-Au-Pd-Pt-Ru is reduced from 335 to 14 compositions with a best-value error of 2.84\%; and Ag-Pd-Pt-Ru and Ag-Pd-Ru retain 31 and 33 candidate compositions, respectively, while preserving the best-performing composition. The least favorable current density result is obtained for Ag-Pd-Pt, where the retained subset contains only 14 of the original 341 candidate compositions and the best-value error remains 16.41\%.

\begin{table*}[htbp]
\centering
\caption{Summary of the literature-guided descriptor-based filtering framework. \textbf{Selected descriptors} gives the two descriptors selected for each performance metric. \textbf{Avg.\ ret.} is the average retained fraction in percent. \textbf{Avg.\ err.} is the average best-value error in percent relative to the full composition search space. Values are reported as mean $\pm$ population standard deviation across the corresponding filtering results. For stability, the averages are calculated from three stability definitions based on different time intervals in the same Fe-Co-Ni composition search space.}
\label{tab:task-summary}
\scriptsize
\setlength{\tabcolsep}{4pt}
\renewcommand{\arraystretch}{1.2}
\begin{ruledtabular}
\begin{tabular}{lccc}
\textbf{Performance metric} & \textbf{Selected descriptors} & \textbf{Avg.\ ret.} & \textbf{Avg.\ err.} \\
Current density & cyclability + rhombic & $6.87 \pm 2.36$ & $3.85 \pm 6.38$ \\
Onset overpotential & instability + spce & $45.45 \pm 0.00$ & $0.00 \pm 0.00$ \\
Resistance & capacity + intermetallic & $27.49 \pm 0.00$ & $0.00 \pm 0.00$ \\
Stability & cyclability + rhombic & $50.00 \pm 0.00$ & $0.00 \pm 0.00$ \\
\end{tabular}
\end{ruledtabular}
\end{table*}

\subsection{Generalization to other performance metrics}

The current density results show that the framework can strongly reduce composition search spaces when descriptors are selected for one performance metric. We next evaluate whether the same descriptor-selection and composition-filtering procedure remains useful when the performance metric changes.

\begin{figure*}[htbp]
\centering
\includegraphics[width=\textwidth]{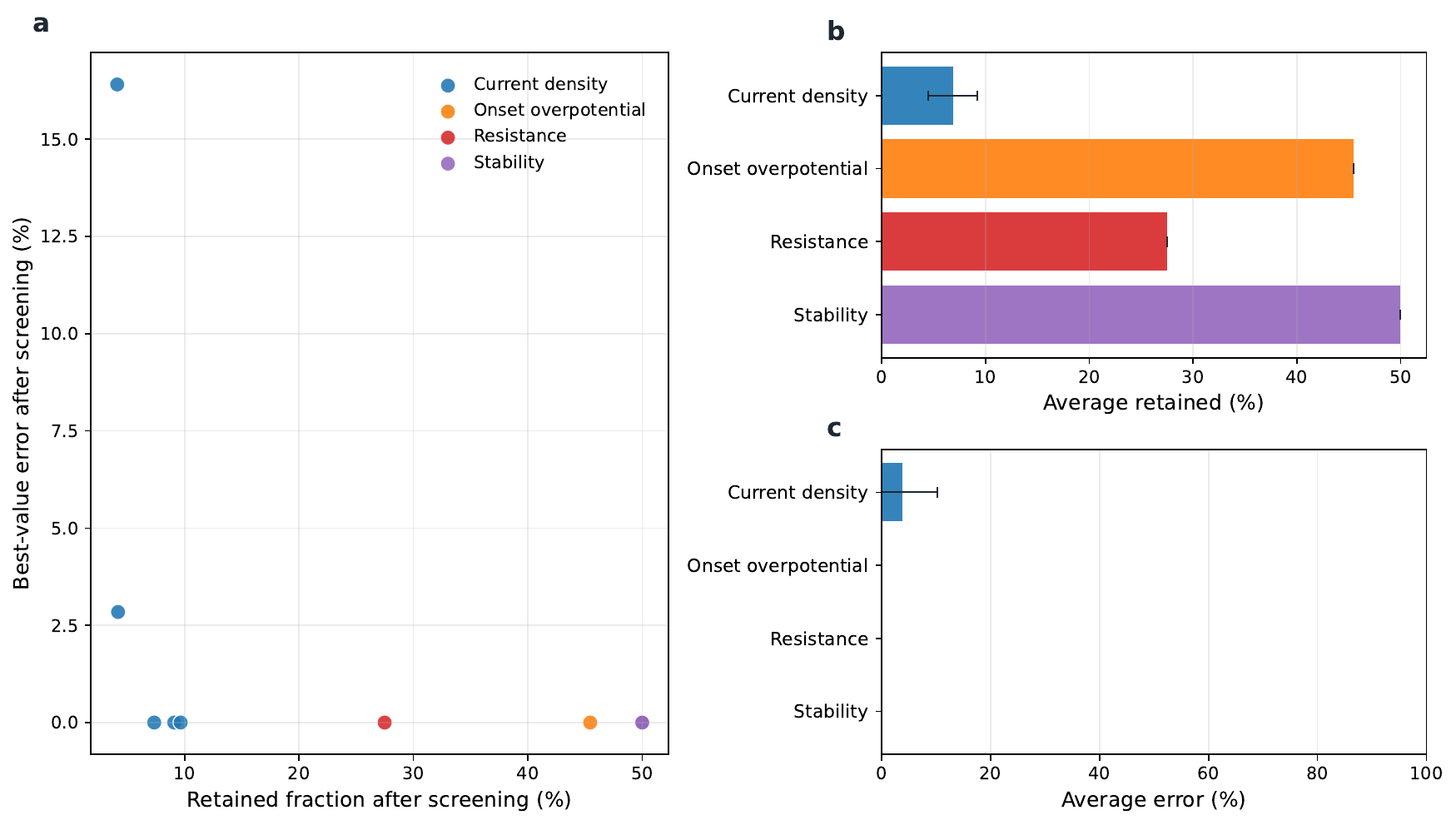}
\caption{Retained-fraction and best-value-error summary. (a) Relation between retained fraction and best-value error, with points colored by performance metric. Each point corresponds to one filtering result. (b) Average retained fraction for each performance metric. (c) Average best-value error for each performance metric on a 0--100\% scale. Error bars in (b) and (c) show population standard deviations across the corresponding filtering results.}
\label{fig:portfolio}
\end{figure*}

In addition to current density, additional performance metrics we evaluate are onset overpotential, resistance, and stability. The results, reported as mean $\pm$ standard deviation where values are aggregated, are summarized in Table~\ref{tab:task-summary}, and detailed filtering results are listed in Tables~\ref{tab:si-onset-overpotential-prediction} - \ref{tab:si-stability-prediction}. Figure~\ref{fig:portfolio} shows all filtering results in a retained-fraction and best-value-error view.

For onset overpotential, the selected descriptors change to \textit{instability} and \textit{spce}. In the Fe-Co-Ni composition search space, the retained subset contains 10 of the original 22 candidate compositions while retaining the best-performing composition. The retained fraction is 45.45\%, and the best-value error is 0.00\%.

For resistance, the selected descriptors are \textit{capacity} and \textit{intermetallic}. In the Ir-Pd-Pt-Rh-Ru composition search space, the retained subset contains 94 of the original 342 candidate compositions and retains the lowest-resistance composition from the full search space. The retained fraction is 27.49\%, and the best-value error is 0.00\%.

For stability, three stability definitions are obtained from the Fe-Co-Ni composition search space by using percentage change over different time intervals. For all three stability definitions, the selected descriptors are again \textit{cyclability} and \textit{rhombic}. The retained subset contains 11 of the original 22 candidate compositions for each time interval, and the best-performing composition is retained in all three stability results. The retained fraction is therefore 50.00\%, and the best-value error remains 0.00\%.

The portfolio view in Figure~\ref{fig:portfolio} shows how retained fraction and best-value error vary across the filtering results for different performance metrics; panels b and c report the corresponding means with standard-deviation error bars. Most points lie close to the horizontal axis, indicating small or zero best-value error. The retained fraction varies more strongly. Composition filtering for current density gives the smallest retained fractions, whereas composition filtering for onset overpotential, resistance, and stability retains broader subsets with zero best-value error. These results show that descriptor selection based on different performance metrics leads to different degrees of composition-space reduction: current density gives a substantial reduction, while onset overpotential, resistance, and stability give larger but still reduced retained subsets that preserve the best-performing compositions.

\subsection{Baseline comparisons}

Our results show that performance metric-dependent descriptor selection can produce compact retained subsets while preserving high-performing compositions. We next compare this selection with two baselines that do not use the performance metric to select descriptors. The first baseline uses the expert-chosen descriptors \textit{dielectric} and \textit{conductivity} from our previous study~\cite{zhang2025electrocatalystdiscoverytextmining} for all performance metrics. This fixed pair provides a useful reference because high-performing electrocatalysts are generally expected to combine favorable electronic and transport behavior across the evaluated metrics. Keeping the same published pair also avoids manual retuning for each task. It is therefore not intended to represent the strongest possible task-specific expert choice. The second baseline uses a 100-seed random-descriptor assessment, where two descriptors are sampled from the shared descriptor vocabulary for each seed and then applied to all filtering results. The expert-chosen baseline is summarized in Table~\ref{tab:expert-baseline-summary}, and the random-descriptor assessment is summarized in Table~\ref{tab:si-random-seed-sweep-summary}.

The expert-chosen baseline shows that fixed descriptors can produce strong reduction, but strong reduction does not always preserve the best-performing composition. For the current density metric, the expert-chosen descriptors reduce the retained fraction to $4.21 \pm 1.04$\%, compared with $6.87 \pm 2.36$\% for the performance metric-dependent descriptors selected by the framework, but the average best-value error increases from $3.85 \pm 6.38$\% to $6.01 \pm 8.83$\%. For resistance, the expert-chosen descriptors reduce the retained fraction to $8.77 \pm 0.00$\%, compared with $27.49 \pm 0.00$\% for the framework-selected descriptors, but introduce a best-value error of $1.67 \pm 0.00$\%, whereas the framework-selected descriptors retain the lowest-resistance composition exactly. For onset overpotential and stability, the expert-chosen descriptors also preserve the best-performing compositions, but retain larger subsets than the framework-selected descriptors.

\begin{table*}[htbp]
\centering
\caption{Summary of the expert-chosen descriptor baseline. The baseline uses \textit{dielectric} and \textit{conductivity} for all evaluated performance metrics. Values are reported as mean $\pm$ population standard deviation across the corresponding filtering results. \textbf{Avg.\ ret.} is the average retained fraction in percent, and \textbf{Avg.\ err.} is the average best-value error in percent relative to the full composition search space. For stability, the averages are calculated from three stability definitions based on different time intervals in the same Fe-Co-Ni composition search space.}
\label{tab:expert-baseline-summary}
\scriptsize
\setlength{\tabcolsep}{4pt}
\renewcommand{\arraystretch}{1.2}
\begin{ruledtabular}
\begin{tabular}{lccc}
\textbf{Performance metric} & \textbf{Expert-chosen descriptors} & \textbf{Avg.\ ret.} & \textbf{Avg.\ err.} \\
Current density & dielectric + conductivity & $4.21 \pm 1.04$ & $6.01 \pm 8.83$ \\
Onset overpotential & dielectric + conductivity & $63.64 \pm 0.00$ & $0.00 \pm 0.00$ \\
Resistance & dielectric + conductivity & $8.77 \pm 0.00$ & $1.67 \pm 0.00$ \\
Stability & dielectric + conductivity & $63.64 \pm 0.00$ & $0.00 \pm 0.00$ \\
\end{tabular}
\end{ruledtabular}
\end{table*}

The random-descriptor assessment shows that arbitrary descriptors can sometimes retain the best-performing composition, but they do not consistently provide the same balance between retained fraction and best-value error as the performance metric-dependent descriptors. For the current density metric, the performance metric-dependent descriptors give a lower retained fraction and a lower best-value error: $6.87 \pm 2.36$\% retained fraction and $3.85 \pm 6.38$\% best-value error, compared with $22.10 \pm 8.54$\% and $5.44 \pm 3.30$\%, respectively, across the random seeds. For onset overpotential, the performance metric-dependent descriptors retain fewer candidate compositions than the random settings, $45.45 \pm 0.00$\% versus $57.00 \pm 16.51$\%, while the random best-value error is $0.76 \pm 3.79$\%. For stability, the performance metric-dependent descriptors give zero best-value error for all three time intervals, whereas the random settings retain $57.00 \pm 16.51$\% of candidates with a best-value error of $218.02 \pm 257.04$\%; the worst random seed reaches 848.22\%. For resistance, the performance metric-dependent descriptors retain more candidate compositions than the random settings, $27.49 \pm 0.00$\% versus $15.28 \pm 10.71$\%, while their respective best-value errors are $0.00 \pm 0.00$\% and $2.57 \pm 4.60$\%.

\section{Discussion}

The central result of this work is that latent knowledge from literature-derived word embeddings can be used to define a reproducible filter for composition-space reduction that is automatically optimized for a given performance metric. Our framework does not only retrieve descriptors that are close to a chosen performance metric. Instead, the descriptor selection itself becomes part of the method: descriptors are selected from a filtered vocabulary in the literature-trained embedding model and are then used to construct a Pareto-based filter for candidate compositions. This changes the role of the embedding model. The embedding model is not only a source of related words, but also a source from which performance metric-dependent descriptors are selected in a systematic, reproducible (deterministic) and performant way.

The framework rests on the empirical hypothesis that suffix-filtered tokens contain literature-derived associations with performance metrics and that two such tokens can provide complementary filtering directions. This hypothesis follows the distributional intuition underlying word embeddings: tokens that share contexts acquire similar representations, although why particular learned associations are useful for a downstream task is not generally available as a formal explanation~\cite{goldberg2014word2vec}. The present results support this hypothesis empirically, but do not establish that the selected suffix patterns or two-descriptor construction are uniquely optimal.

This distinction is important compared with approaches that rely on expert-chosen descriptors~\cite{zhang2025electrocatalystdiscoverytextmining}. Descriptors such as \textit{dielectric} and \textit{conductivity} can be chemically meaningful and can produce useful filtering behavior, but the descriptor choice is fixed before the filtering begins. The reasoning behind the choice therefore remains partly outside the method. In our framework, the descriptor choice is made from the same literature-derived representation that encodes the contextual relations between performance metrics, descriptors, and materials-related language. The framework therefore provides a route from latent literature knowledge to an interpretable and reproducible composition-space filter.

The descriptor-selection step is important because the selected descriptors determine which candidate compositions are favored by the filter. A nearest-neighbor strategy would mainly recover descriptors that are already close to the performance metric, but such descriptors may be redundant or may just not provide a useful contrast for filtering. By also considering the cosine similarity between the two descriptors, the Pareto-based selection favors descriptor combinations that balance association with the performance metric and contrast between descriptors. The ranking by similarity to the word \textit{performance} further keeps the selected descriptors associated with general performance-related language in the literature. The advantage is that the filter is not based on a single broad descriptor or on two nearly identical descriptors. Instead, the filter is defined by two interpretable descriptors that provide complementary similarity views of the candidate compositions.

The selected descriptors are performance metric-dependent, but not arbitrary. \textit{Cyclability} and \textit{rhombic} are selected for current density and recur for stability, whereas onset overpotential and resistance lead to \textit{instability} and \textit{spce}, and \textit{capacity} and \textit{intermetallic}, respectively. This result is useful because descriptor selection changes when the performance metric changes, but does not change randomly for every metric. If the same descriptors were selected for all performance metrics, the selection step would appear insensitive to the metric being evaluated. If completely unrelated descriptors were selected for every metric, the selection could be interpreted as unstable, i.e., not deterministic. The observed behavior instead suggests that the embedding model contains performance metric-dependent associations that can change across performance metrics while still allowing recurrence when similar literature-derived signals are present.

After descriptor selection, the two selected descriptors provide the link between the literature-derived embedding model and the composition search space. Each candidate composition, represented by a normalized vector generated from its elemental fractions, is projected onto the two selected descriptors by calculating two similarity values: one similarity to the first descriptor and one similarity to the second descriptor. Pareto filtering is then applied in the resulting two-dimensional descriptor-similarity projection space. Each point in this projection space corresponds to one candidate composition and the non-dominated points directly define the retained subset. This gives the selected descriptors an operational role: they are not only interpretable descriptors from the literature-derived vocabulary, but also define how the composition search space is narrowed. Our proposed framework uses Pareto filtering in two different spaces for two different purposes: first in the literature-derived embedding space to select descriptors, and then in the descriptor-similarity projection space to retain candidate compositions. The method does not extract explicit facts from individual sentences, and it does not replace the quantitative, measured performance metric with a direct text-based predictor. Instead, it uses literature-derived embedding associations to construct a filter that narrows the candidate composition space.

The practical value of the framework appears in the retained fractions and best-value errors. For the current density metric, the framework reduces the candidate compositions to an average retained fraction of $6.87 \pm 2.36$\%, while keeping the average best-value error at $3.85 \pm 6.38$\%. For onset overpotential, resistance, and stability, the retained fractions are larger, but the best-value error remains $0.00 \pm 0.00$\% in the evaluated results. This variation shows that the framework does not yield one fixed retained fraction. Some performance metrics lead to a highly selective retained subset, whereas others lead to broader retained subsets that still preserve the best-performing composition. The useful outcome is therefore not simply the smallest possible retained subset, but a retained subset that reduces the composition search space while keeping the optimum or a near-optimum composition.

The two baseline comparisons clarify the value of performance metric-dependent descriptor selection. The expert-chosen descriptors \textit{dielectric} and \textit{conductivity} can reduce the composition search space strongly, but stronger reduction can come with larger best-value error, as seen for current density and resistance. The random-descriptor assessment provides a broader comparison with arbitrary descriptor choices. Random descriptors can sometimes retain the best-performing composition, but the filtering outcome varies strongly across seeds and performance metrics, especially for \textit{stability}. The performance metric-dependent descriptors selected by our framework are therefore not claimed to be unique. Rather, the results show that framework-selected descriptors give a more controlled balance between retained fraction and best-value error than descriptors fixed in advance or sampled at random.

The present study also has clear limitations. Current density is evaluated across five composition search spaces, while onset overpotential and resistance are each evaluated on one composition search space. Stability is evaluated using three definitions based on different time intervals, but all three stability definitions come from the same Fe-Co-Ni composition search space.
Broader validation across larger composition search spaces, additional performance metrics, different literature corpora, and different embedding constructions is needed to test how stable the framework-selected descriptors remain when the literature source or the candidate compositions change.
However, this would require access to more high-quality, multimodal datasets. The limited number of such datasets reflects a broader constraint in materials research: dense experimental datasets are often small because they are expensive to obtain.
The descriptor \textit{spce} also shows that vocabulary filtering deserves further attention, because corpus-specific or ambiguous descriptors can enter the filtering procedure. Even with these limitations, our results show that the scientific literature can support composition-space filtering in a different way than just a simple descriptor lookup: literature-derived embeddings can provide interpretable, automatically selected descriptors for narrowing candidate composition spaces while preserving high-performing compositions.

\section{Conclusion}

We present a literature-guided descriptor-based framework for filtering composition search spaces with respect to high-performing candidate compositions. For each performance metric, the framework selects two descriptors from a prefiltered vocabulary in a literature-trained word embedding space and uses the selected descriptors to construct a Pareto-based filter for candidate compositions. Across the evaluated performance metrics, the framework reduces the candidate composition spaces while preserving the best or near-best measured compositions. The framework-selected descriptors also vary with the performance metric, with \textit{cyclability} and \textit{rhombic} recurring for current density and stability. Compared with expert-chosen and random descriptors, our performance metric-dependent descriptors provide a reproducible, controlled balance between retained fraction and best-value error. Our results show that literature-derived embeddings can support interpretable and reproducible composition-space filters for narrowing candidate compositions.

\section{Methods}
\label{sec:methods}

\begin{figure*}[htbp]
\centering
\includegraphics[width=\textwidth]{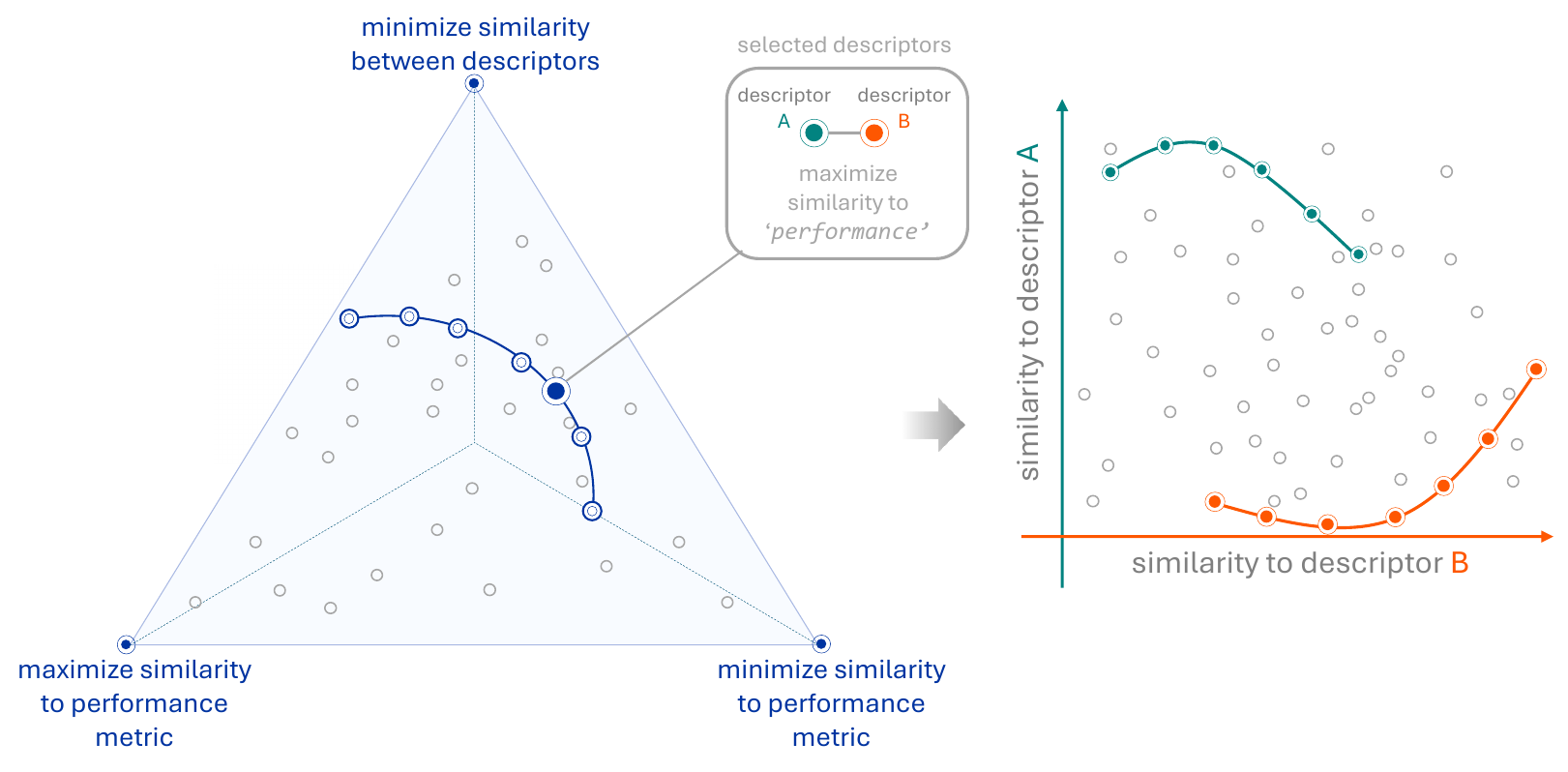}
\caption{Schematic of the literature-guided descriptor-based filtering framework.}
\label{fig:workflow}
\end{figure*}

The framework follows a two-part route from literature-derived embeddings to composition-space filtering, as summarized in Figure~\ref{fig:workflow}. First, a performance metric is represented in the literature-trained word embedding model, and descriptors are selected from a filtered descriptor vocabulary. All possible two-descriptor combinations are evaluated in an embedding-space selection problem that balances three objectives: one descriptor should remain close to the performance metric, the other should provide contrast in its relation to the performance metric, and the two descriptors should not be too similar to each other. The resulting Pareto front can contain several descriptor combinations, so the final combination is chosen by the average similarity of the two descriptors to the word \textit{performance}. Second, each candidate composition is projected onto the two selected descriptors by computing its cosine similarity to each descriptor. This creates a two-dimensional descriptor-similarity projection space in which Pareto filtering is applied. The output is a retained subset of candidate compositions, together with the retained fraction, the best-value error, and whether the best-performing composition of the full composition search space is retained.

\subsection{Word embeddings and performance metric representation}

All embedding-space operations use a pretrained Word2Vec model trained on a domain-specific scientific corpus~\cite{Zhang2024}. Let $\mathbf{v}(w)\in\mathbb{R}^{d}$, with $d=200$, denote the $\ell_{2}$-normalized embedding vector of vocabulary word $w$. For a performance metric written as text $t$, the corresponding embedding vector $\mathbf{v}_{t}$ is defined directly from the model vocabulary when $t$ appears as a single vocabulary word. When the performance metric is expressed as a multi-word phrase outside the vocabulary, the phrase is split on spaces, underscores, hyphens, and slashes, and $\mathbf{v}_{t}$ is defined as the normalized average of all in-vocabulary component words,
\begin{equation}
\mathbf{v}_{t}=
\begin{cases}
\mathbf{v}(t), & \text{if } t \text{ is in the vocabulary},\\[4pt]
\mathrm{norm}\!\left(\dfrac{1}{|T|}\sum\limits_{u\in T}\mathbf{v}(u)\right), & \text{otherwise},
\end{cases}
\end{equation}
where $T$ is the set of valid in-vocabulary component words. This representation allows single-word and multi-word performance metrics to be treated in the same embedding space as the descriptors.

\subsection{Descriptor vocabulary}

Descriptors are selected from a filtered descriptor vocabulary derived from the same embedding-model vocabulary used to represent the performance metric. Here, a token becomes a candidate descriptor when it passes the vocabulary filter, and its 200-dimensional embedding vector provides its numerical representation. A candidate descriptor becomes a selected descriptor when it is chosen for filtering candidate compositions for a specific performance metric. A vocabulary word is retained as a candidate descriptor only if it is at least four characters long, contains alphabetic characters only, appears in lowercase form, and ends with at least one suffix pattern from the set
\begin{equation}
\mathcal{S}=\{\texttt{*ivity},\texttt{*able},\texttt{*ive},\texttt{*ous},\texttt{*ic},\texttt{*ity},\texttt{*ce}\}.
\end{equation}
Here, \texttt{*} denotes any preceding character sequence, so the patterns define suffix-based vocabulary filtering. The resulting descriptor vocabulary is denoted by
\begin{equation}
\mathcal{W}=\{w_{1},w_{2},\dots,w_{M}\}.
\end{equation}
For each descriptor $w_{i}$, its similarity to the performance-metric vector is computed as
\begin{equation}
s_{i}=\cos\!\bigl(\mathbf{v}(w_{i}),\mathbf{v}_{t}\bigr).
\end{equation}

\subsection{Descriptor selection}

All unordered two-descriptor combinations $(w_{i},w_{j})$ with $i<j$ are evaluated from the descriptor vocabulary. Two descriptors are used as the smallest representation that supports a trade-off-based Pareto filter and a directly interpretable two-dimensional projection. This is an exploratory design choice rather than a claim that two descriptors are uniquely optimal. For each combination, three quantities are calculated,
\begin{align}
s_{ij}^{\min} &= \min(s_{i},s_{j}),\\
s_{ij}^{\max} &= \max(s_{i},s_{j}),\\
c_{ij} &= \cos\!\bigl(\mathbf{v}(w_{i}),\mathbf{v}(w_{j})\bigr).
\end{align}
The descriptor combinations are filtered by a three-objective Pareto selection that minimizes $s_{ij}^{\min}$, maximizes $s_{ij}^{\max}$, and minimizes $c_{ij}$. This formulation tests the hypothesis that one descriptor should remain relatively close to the performance metric while the second provides contrast by representing a different aspect of the embedding space. Maximizing $s_{ij}^{\max}$ promotes the first role, whereas minimizing $s_{ij}^{\min}$ and $c_{ij}$ promotes the second and discourages redundant descriptor pairs. The corresponding non-dominated combinations define the descriptor-level Pareto front.

The Pareto front can contain multiple descriptor combinations. To select one final combination, each descriptor combination on the Pareto front is ranked by its average similarity to the word \textit{performance}. Let $\mathbf{v}_{\mathrm{perf}}$ denote the embedding vector of \textit{performance}. The ranking score is
\begin{equation}
r_{ij}=\frac{1}{2}\left[\cos\!\bigl(\mathbf{v}(w_{i}),\mathbf{v}_{\mathrm{perf}}\bigr)+\cos\!\bigl(\mathbf{v}(w_{j}),\mathbf{v}_{\mathrm{perf}}\bigr)\right].
\end{equation}
The descriptor combination with the largest $r_{ij}$ is selected for composition-space filtering.

\subsection{Composition representation and descriptor-similarity projection}

In this work, each composition search space is represented as a table containing elemental fractions in columns named by periodic-table symbols, together with the measured values of the relevant performance metric. The measured performance values are not used during filtering; they are used only afterward to evaluate the retained subset. Composition vectors are generated with the MatNexus \texttt{MaterialSimilarityCalculator}~\cite{Zhang2024}, yielding a normalized vector $\mathbf{m}_{n}$ for each candidate composition $n$.

For the selected descriptors $(w_{i},w_{j})$, each candidate composition is projected into a two-dimensional descriptor-similarity space by computing its similarity to both descriptor embeddings,
\begin{align}
a_{n} &= \cos \bigl(\mathbf{m}_{n},\mathbf{v}(w_{i})\bigr),\\
b_{n} &= \cos \bigl(\mathbf{m}_{n},\mathbf{v}(w_{j})\bigr).
\end{align}
Thus, each candidate composition is represented by the point $(a_{n},b_{n})$ in descriptor-similarity projection space.

Candidate compositions are retained using the bidirectional two-objective Pareto filtering strategy adapted from~\cite{zhang2025electrocatalystdiscoverytextmining}. A candidate composition is retained if its point belongs to either of two two-objective Pareto fronts: one obtained by maximizing $a_{n}$ while minimizing $b_{n}$, and the other obtained by minimizing $a_{n}$ while maximizing $b_{n}$. The union of these two fronts defines the retained subset $\mathcal{P}_{ij}$ for the selected descriptors $(w_{i},w_{j})$.

\subsection{Evaluation metrics}

Let $N$ denote the number of candidate compositions in the full composition search space, and let $y_{n}$ denote the measured value of the performance metric for candidate composition $n$. The retained fraction for selected descriptors $(w_{i},w_{j})$ is defined as
\begin{equation}
R_{ij}=100\times\frac{|\mathcal{P}_{ij}|}{N}.
\end{equation}
Lower values of $R_{ij}$ indicate stronger reduction of the composition search space.

To quantify preservation of the optimum, let $y^{\ast}$ denote the best value in the full composition search space and let $y_{ij}^{\ast}$ denote the best value inside the retained subset. For minimization metrics,
\begin{equation}
y^{\ast}=\min_{n}y_{n}, \qquad y_{ij}^{\ast}=\min_{n\in\mathcal{P}_{ij}}y_{n},
\end{equation}
whereas for maximization metrics,
\begin{equation}
y^{\ast}=\max_{n}y_{n}, \qquad y_{ij}^{\ast}=\max_{n\in\mathcal{P}_{ij}}y_{n}.
\end{equation}
The best-value error is defined as
\begin{equation}
E_{ij}=100\times\frac{|y_{ij}^{\ast}-y^{\ast}|}{|y^{\ast}|}.
\end{equation}
In addition to $R_{ij}$ and $E_{ij}$, the analysis records whether the best-performing composition from the full composition search space is retained and the rank of the best retained composition within the full search space.

\subsection{Baseline settings}

To assess the effect of performance metric-dependent descriptor selection, two descriptor baselines are evaluated by bypassing the descriptor-selection step and supplying descriptors directly to the composition-filtering step. The first baseline uses the expert-chosen descriptors \textit{dielectric} and \textit{conductivity} for all evaluated performance metrics. The second baseline is a 100-seed random-descriptor assessment. For this assessment, the candidate pool is defined as the sorted intersection of the suffix-filtered descriptor vocabularies from the evaluated performance metrics. Seeds 0 through 99 are evaluated, with two descriptors sampled per seed from this shared descriptor vocabulary and then applied to the evaluated filtering results.

In both baseline settings, the composition-filtering procedure remains unchanged. The supplied descriptors are evaluated with the same composition representation, the same descriptor-similarity projection, the same bidirectional two-objective Pareto filtering, and the same retained-fraction and best-value-error metrics. The baseline comparison therefore isolates the effect of selecting descriptors in a performance metric-dependent way rather than fixing descriptors in advance or sampling descriptors at random.

\begin{acknowledgments}
LZ and MS acknowledge funding by the Deutsche Forschungsgemeinschaft (DFG, German Research Foundation) under CRC 1625, project number 506711657 and LZ funding by the DFG through TRR 247, project number 388390466.
\end{acknowledgments}

\section*{Author contributions}
\textbf{Lei Zhang:} Conceptualization, Methodology, Software, Validation, Formal analysis, Investigation, Data Curation, Writing – Original Draft, Visualization, Experimentation\\
\textbf{Markus Stricker:} Conceptualization, Resources, Supervision, Writing – Review \& Editing, Funding acquisition. 

\section*{Competing interests}
The authors declare no competing interests.

\section*{Data availability}

The datasets used for the current-density prediction task are available from Zenodo at \url{https://doi.org/10.5281/zenodo.13992986} and \url{https://doi.org/10.5281/zenodo.14959252}. The dataset used for the resistance prediction task is available from Zenodo at \url{https://doi.org/10.5281/zenodo.7249698}. The onset-overpotential and stability data are taken from the study reported at \url{https://doi.org/10.1021/acsmaterialsau.2c00016}. The code used to perform the descriptor-pair discovery, Pareto candidate reduction, and baseline analyses is available at \url{https://github.com/lab-mids/w2v_descriptor_pairs}.

\bibliographystyle{apsrev4-2}
\bibliography{references}

@inproceedings{mikolov2013phrases,
  author = {Mikolov, Tomas and Sutskever, Ilya and Chen, Kai and Corrado, Greg S and Dean, Jeff},
  title = {Distributed Representations of Words and Phrases and their Compositionality},
  booktitle = {Advances in Neural Information Processing Systems 26},
  editor = {Burges, C. J. C. and Bottou, L. and Welling, M. and Ghahramani, Z. and Weinberger, K. Q.},
  publisher = {Curran Associates, Inc.},
  year = {2013},
  url = {https://proceedings.neurips.cc/paper_files/paper/2013/file/9aa42b31882ec039965f3c4923ce901b-Paper.pdf}
}

@article{tshitoyan2019materials,
  author = {Tshitoyan, Vahe and Dagdelen, John and Weston, Leigh and Dunn, Alexander and Rong, Ziqin and Kononova, Olga and Persson, Kristin A. and Ceder, Gerbrand and Jain, Anubhav},
  title = {Unsupervised word embeddings capture latent knowledge from materials science literature},
  journal = {Nature},
  volume = {571},
  number = {7763},
  pages = {95--98},
  year = {2019},
  month = jul,
  doi = {10.1038/s41586-019-1335-8},
  url = {https://doi.org/10.1038/s41586-019-1335-8}
}

@article{qu2024language,
  author = {Qu, Jiaxing and Xie, Yuxuan Richard and Ciesielski, Kamil M. and Porter, Claire E. and Toberer, Eric S. and Ertekin, Elif},
  title = {Leveraging language representation for materials exploration and discovery},
  journal = {npj Computational Materials},
  volume = {10},
  number = {1},
  pages = {58},
  year = {2024},
  month = mar,
  doi = {10.1038/s41524-024-01231-8},
  url = {https://doi.org/10.1038/s41524-024-01231-8}
}

@article{jiang2025nlp,
  author = {Jiang, Xue and Wang, Weiren and Tian, Shaohan and Wang, Hao and Lookman, Turab and Su, Yanjing},
  title = {Applications of natural language processing and large language models in materials discovery},
  journal = {npj Computational Materials},
  volume = {11},
  number = {1},
  pages = {79},
  year = {2025},
  month = mar,
  doi = {10.1038/s41524-025-01554-0},
  url = {https://doi.org/10.1038/s41524-025-01554-0}
}

@Article{Ludwig2019,
  author    = {Ludwig, Alfred},
  title     = {Discovery of new materials using combinatorial synthesis and high-throughput characterization of thin-film materials libraries combined with computational methods},
  journal   = {npj Computational Materials},
  year      = {2019},
  volume    = {5},
  number    = {1},
  pages     = {70},
  doi       = {https://doi.org/10.1038/s41524-019-0205-0},
  publisher = {Nature Publishing Group},
}

@article{butler2018machine,

  title   = {Machine learning for molecular and materials science},

  author  = {Butler, Keith T. and Davies, Daniel W. and Cartwright, Hugh and Isayev, Olexandr and Walsh, Aron},

  journal = {Nature},

  volume  = {559},

  number  = {7715},

  pages   = {547--555},

  year    = {2018},

  doi     = {10.1038/s41586-018-0337-2}

}

@article{jain2013materialsproject,

  title   = {Commentary: The Materials Project: A materials genome approach to accelerating materials innovation},

  author  = {Jain, Anubhav and Ong, Shyue Ping and Hautier, Geoffroy and Chen, Wei and Richards, William Davidson and Dacek, Stephen and Cholia, Shreyas and Gunter, Dan and Skinner, David and Ceder, Gerbrand and Persson, Kristin A.},

  journal = {APL Materials},

  volume  = {1},

  number  = {1},

  pages   = {011002},

  year    = {2013},

  doi     = {10.1063/1.4812323}

}

@article{zakutayev2017highthroughput,

  title   = {Fulfilling the promise of the materials genome initiative with high-throughput experimental methodologies},

  author  = {Zakutayev, Andriy and Wunder, Nick and Schwarting, Marcus and Perkins, John D. and White, Richard and Munch, Kristin and Tumas, William and Phillips, Caleb},

  journal = {Applied Physics Reviews},

  volume  = {4},

  number  = {1},

  pages   = {011105},

  year    = {2017},

  doi     = {10.1063/1.4977487}

}

@article{chen2025accelerating,

  title   = {Accelerating computational materials discovery with artificial intelligence and cloud high-performance computing: from large-scale screening to experimental validation},

  author  = {Chen, Chi and Nguyen, Dan Thien and Lee, Shannon J. and Baker, Nathan A. and Karakoti, Ajay S. and Lauw, Linda and Owen, Craig and Mueller, Karl T. and Bilodeau, Brian A. and Murugesan, Vijayakumar and Troyer, Matthias},

  journal = {npj Computational Materials},

  volume  = {11},

  pages   = {73},

  year    = {2025},

  doi     = {10.1038/s41524-025-01570-0}

}

@article{curtarolo2012aflow,

  title   = {AFLOW: An automatic framework for high-throughput materials discovery},

  author  = {Curtarolo, Stefano and Setyawan, Wahyu and Hart, Gus L. W. and Jahnatek, Michal and Chepulskii, Roman V. and Taylor, Richard H. and Wang, Shidong and Xue, Junkai and Yang, Kesong and Levy, Ohad and Mehl, Michael J. and Stokes, Harold T. and Demchenko, Denis O. and Morgan, Dane},

  journal = {Computational Materials Science},

  volume  = {58},

  pages   = {218--226},

  year    = {2012},

  doi     = {10.1016/j.commatsci.2012.02.005}

}

@misc{zhang2025electrocatalystdiscoverytextmining,
      title={Electrocatalyst discovery through text mining and multi-objective optimization}, 
      author={Lei Zhang and Markus Stricker},
      year={2025},
      eprint={2502.20860},
      archivePrefix={arXiv},
      primaryClass={cond-mat.mtrl-sci},
      url={https://arxiv.org/abs/2502.20860}, 
}

@article{zhang2025compositionproperty,
      author={Zhang, Lei and Banko, Lars and Schuhmann, Wolfgang and Ludwig, Alfred and Stricker, Markus},
      title={Composition-property extrapolation for compositionally complex solid solutions based on word embeddings},
      journal={Digital Discovery},
      year={2025},
      volume={4},
      pages={1578--1590},
      doi={10.1039/D5DD00169B},
}

@ARTICLE{Zhang2024,
	author = {Zhang, Lei and Stricker, Markus},
	title = {MATNEXUS: A comprehensive text mining and analysis suite for materials discovery},
	year = {2024},
	journal = {SoftwareX},
	volume = {26},
	doi = {10.1016/j.softx.2024.101654},
    pages = {101654},

}

@misc{goldberg2014word2vec,
      title={word2vec Explained: deriving Mikolov et al.'s negative-sampling word-embedding method},
      author={Goldberg, Yoav and Levy, Omer},
      year={2014},
      eprint={1402.3722},
      archivePrefix={arXiv},
      primaryClass={cs.CL},
      doi={10.48550/arXiv.1402.3722},
}

\clearpage
\onecolumngrid
\appendix
\setcounter{figure}{0}
\renewcommand{\thefigure}{A\arabic{figure}}

\setcounter{table}{0}
\renewcommand{\thetable}{A\arabic{table}}

\section{Supplementary Information}

\section*{Supplementary Tables}

\begin{table}[!htbp]
    \centering
\caption{Detailed composition-filtering results for the current density performance metric. The applied potential identifies the current-density measurement used for each composition search space. Full count and retained count give the number of candidate compositions before and after Pareto filtering. Best/full and best/retained report the best current-density values in mA cm$^{-2}$, and best-value error is reported relative to the full composition search space.}

    \label{tab:si-current-density-prediction}
    \scriptsize
    \setlength{\tabcolsep}{5pt}
    \resizebox{\textwidth}{!}{%
    \begin{tabular}{lllrrrrr}
        \toprule
        Composition search space & Applied potential & Selected descriptors & Full count & Retained count & Best/full & Best/retained & Best-value error (\%) \\
        \midrule
        Ag-Au-Pd-Pt-Rh & -300 mV & cyclability + rhombic & 327 & 24 & -1.13 & -1.13 & 0.00 \\
        Ag-Au-Pd-Pt-Ru & -300 mV & cyclability + rhombic & 335 & 14 & -1.49 & -1.45 & 2.84 \\
        Ag-Pd-Pt & 850 mV & cyclability + rhombic & 341 & 14 & -0.58 & -0.49 & 16.41 \\
        Ag-Pd-Pt-Ru & 850 mV & cyclability + rhombic & 341 & 31 & -0.37 & -0.37 & 0.00 \\
        Ag-Pd-Ru & 850 mV & cyclability + rhombic & 342 & 33 & -0.67 & -0.67 & 0.00 \\
        \bottomrule
    \end{tabular}%
    }
\end{table}

\begin{table}[!htbp]
    \centering
\caption{Detailed composition-filtering result for the onset overpotential performance metric. Full count and retained count give the number of candidate compositions before and after Pareto filtering. Best/full and best/retained are reported in mV, and best-value error is reported relative to the full composition search space.}
    \label{tab:si-onset-overpotential-prediction}
    \scriptsize
    \setlength{\tabcolsep}{5pt}
    \resizebox{\textwidth}{!}{%
    \begin{tabular}{lllrrrr}
        \toprule
        Composition search space & Selected descriptors & Full count & Retained count & Best/full & Best/retained & Best-value error (\%) \\
        \midrule
        Fe-Co-Ni & instability + spce & 22 & 10 & 394.00 & 394.00 & 0.00 \\
        \bottomrule
    \end{tabular}%
    }
\end{table}

\begin{table}[!htbp]
    \centering
\caption{Detailed composition-filtering result for the resistance performance metric. Full count and retained count give the number of candidate compositions before and after Pareto filtering. Best/full and best/retained report the lowest-resistance values, and best-value error is reported relative to the full composition search space.}
    \label{tab:si-resistance-prediction}
    \scriptsize
    \setlength{\tabcolsep}{5pt}
    \resizebox{\textwidth}{!}{%
    \begin{tabular}{lllrrrr}
        \toprule
        Composition search space & Selected descriptors & Full count & Retained count & Best/full & Best/retained & Best-value error (\%) \\
        \midrule
        Ir-Pd-Pt-Rh-Ru & capacity + intermetallic & 342 & 94 & 1.49 & 1.49 & 0.00 \\
        \bottomrule
    \end{tabular}%
    }
\end{table}

\begin{table}[!htbp]
    \centering
\caption{Detailed composition-filtering results for the stability performance metric. All three rows come from the same Fe-Co-Ni composition search space but use different stability definitions based on different time intervals. Here $e_0$ is the onset overpotential in mV, $e_1$ is the overpotential after 1 h, and $e_2$ is the overpotential after 2 h. The time interval gives the percentage-change interval used to define stability. Full count and retained count give the number of candidate compositions before and after Pareto filtering. Best/full and best/retained are reported as percentage change, where smaller values indicate higher stability.}
    \label{tab:si-stability-prediction}
    \scriptsize
    \setlength{\tabcolsep}{5pt}
    \resizebox{\textwidth}{!}{%
    \begin{tabular}{lllrrrrr}
        \toprule
        Composition search space & Time interval & Selected descriptors & Full count & Retained count & Best/full & Best/retained & Best-value error (\%) \\
        \midrule
        Fe-Co-Ni  & $e_0 \to e_1$ & cyclability + rhombic & 22 & 11 & 0.59 & 0.59 & 0.00 \\
        Fe-Co-Ni &  $e_1 \to e_2$ & cyclability + rhombic & 22 & 11 & 0.20 & 0.20 & 0.00 \\
        Fe-Co-Ni &  $e_0 \to e_2$ & cyclability + rhombic & 22 & 11 & 0.40 & 0.40 & 0.00 \\
        \bottomrule
    \end{tabular}%
    }
\end{table}

\begin{table}[!htbp]
    \centering
\caption{Detailed composition-filtering results for the current density performance metric using the expert-chosen descriptor baseline \textit{dielectric + conductivity}. The applied potential identifies the current-density measurement used for each composition search space. Full count and retained count give the number of candidate compositions before and after Pareto filtering. Best/full and best/retained report the best current-density values in mA cm$^{-2}$, and best-value error is reported relative to the full composition search space.}
    \label{tab:si-baseline-current-density-prediction}
    \scriptsize
    \setlength{\tabcolsep}{5pt}
    \resizebox{\textwidth}{!}{%
    \begin{tabular}{lllrrrrr}
        \toprule
        Composition search space & Applied potential & Expert-chosen descriptors & Full count & Retained count & Best/full & Best/retained & Best-value error (\%) \\
        \midrule
        Ag-Au-Pd-Pt-Rh & -300 mV & dielectric + conductivity & 327 & 9 & -1.13 & -1.09 & 3.80 \\
        Ag-Au-Pd-Pt-Ru & -300 mV & dielectric + conductivity & 335 & 20 & -1.49 & -1.45 & 2.84 \\
        Ag-Pd-Pt & 850 mV & dielectric + conductivity & 341 & 15 & -0.58 & -0.45 & 23.40 \\
        Ag-Pd-Pt-Ru & 850 mV & dielectric + conductivity & 341 & 13 & -0.37 & -0.37 & 0.00 \\
        Ag-Pd-Ru & 850 mV & dielectric + conductivity & 342 & 14 & -0.67 & -0.67 & 0.00 \\
        \bottomrule
    \end{tabular}%
    }
\end{table}

\begin{table}[!htbp]
    \centering
\caption{Detailed composition-filtering result for the onset overpotential performance metric using the expert-chosen descriptor baseline \textit{dielectric + conductivity}. Full count and retained count give the number of candidate compositions before and after Pareto filtering. Best/full and best/retained are reported in mV, and best-value error is reported relative to the full composition search space.}
    \label{tab:si-baseline-onset-overpotential-prediction}
    \scriptsize
    \setlength{\tabcolsep}{5pt}
    \resizebox{\textwidth}{!}{%
    \begin{tabular}{lllrrrr}
        \toprule
        Composition search space & Expert-chosen descriptors & Full count & Retained count & Best/full & Best/retained & Best-value error (\%) \\
        \midrule
        Fe-Co-Ni & dielectric + conductivity & 22 & 14 & 394.00 & 394.00 & 0.00 \\
        \bottomrule
    \end{tabular}%
    }
\end{table}

\begin{table}[!htbp]
    \centering
\caption{Detailed composition-filtering result for the resistance performance metric using the expert-chosen descriptor baseline \textit{dielectric + conductivity}. Full count and retained count give the number of candidate compositions before and after Pareto filtering. Best/full and best/retained report the lowest-resistance values, and best-value error is reported relative to the full composition search space.}
    \label{tab:si-baseline-resistance-prediction}
    \scriptsize
    \setlength{\tabcolsep}{5pt}
    \resizebox{\textwidth}{!}{%
    \begin{tabular}{lllrrrr}
        \toprule
        Composition search space & Expert-chosen descriptors & Full count & Retained count & Best/full & Best/retained & Best-value error (\%) \\
        \midrule
        Ir-Pd-Pt-Rh-Ru & dielectric + conductivity & 342 & 30 & 1.49 & 1.47 & 1.67 \\
        \bottomrule
    \end{tabular}%
    }
\end{table}

\begin{table}[!htbp]
    \centering
\caption{Detailed composition-filtering results for the stability performance metric using the expert-chosen descriptor baseline \textit{dielectric + conductivity}. All three rows come from the same Fe-Co-Ni composition search space but use different stability definitions based on different time intervals. Here $e_0$ is the onset overpotential in mV, $e_1$ is the overpotential after 1 h, and $e_2$ is the overpotential after 2 h. The time interval gives the percentage-change interval used to define stability. Full count and retained count give the number of candidate compositions before and after Pareto filtering. Best/full and best/retained are reported as percentage change, where smaller values indicate higher stability.}
    \label{tab:si-baseline-stability-prediction}
    \scriptsize
    \setlength{\tabcolsep}{5pt}
    \resizebox{\textwidth}{!}{%
    \begin{tabular}{lllrrrrr}
        \toprule
        Composition search space  & Time interval & Expert-chosen descriptors & Full count & Retained count & Best/full & Best/retained & Best-value error (\%) \\
        \midrule
        Fe-Co-Ni  & $e_0 \to e_1$ & dielectric + conductivity & 22 & 14 & 0.59 & 0.59 & 0.00 \\
        Fe-Co-Ni & $e_1 \to e_2$ & dielectric + conductivity & 22 & 14 & 0.20 & 0.20 & 0.00 \\
        Fe-Co-Ni & $e_0 \to e_2$ & dielectric + conductivity & 22 & 14 & 0.40 & 0.40 & 0.00 \\
        \bottomrule
    \end{tabular}%
    }
\end{table}

\begin{table}[!htbp]
    \centering
\caption{Summary of the 100-seed random-descriptor assessment. For each seed, two descriptors were sampled from the shared descriptor vocabulary and then applied to all evaluated filtering results. Reported values are performance-metric-level summaries across the corresponding composition search spaces. Standard deviations are population standard deviations across the 100 seeds. Because the stability metric is a percentage change with best values close to zero, unsuitable random descriptors can produce very large best-value errors for that metric.}
    \label{tab:si-random-seed-sweep-summary}
    \scriptsize
    \setlength{\tabcolsep}{4pt}
    \resizebox{\textwidth}{!}{%
    \begin{tabular}{lrrrrrrr}
        \toprule
        Performance metric & Seeds & Mean ret. (\%) & SD ret. (\%) & Mean err. (\%) & SD err. (\%) & Best err. (\%) & Worst err. (\%) \\
        \midrule
        Current density & 100 & 22.10 & 8.54 & 5.44 & 3.30 & 0.09 & 19.18 \\
        Onset overpotential & 100 & 57.00 & 16.51 & 0.76 & 3.79 & 0.00 & 22.08 \\
        Resistance & 100 & 15.28 & 10.71 & 2.57 & 4.60 & 0.00 & 20.24 \\
        Stability & 100 & 57.00 & 16.51 & 218.02 & 257.04 & 0.00 & 848.22 \\
        \bottomrule
    \end{tabular}%
    }
\end{table}

\clearpage
\section*{Supplementary Figures}

\begin{figure}[!htbp]
    \centering
    \includegraphics[width=\textwidth]{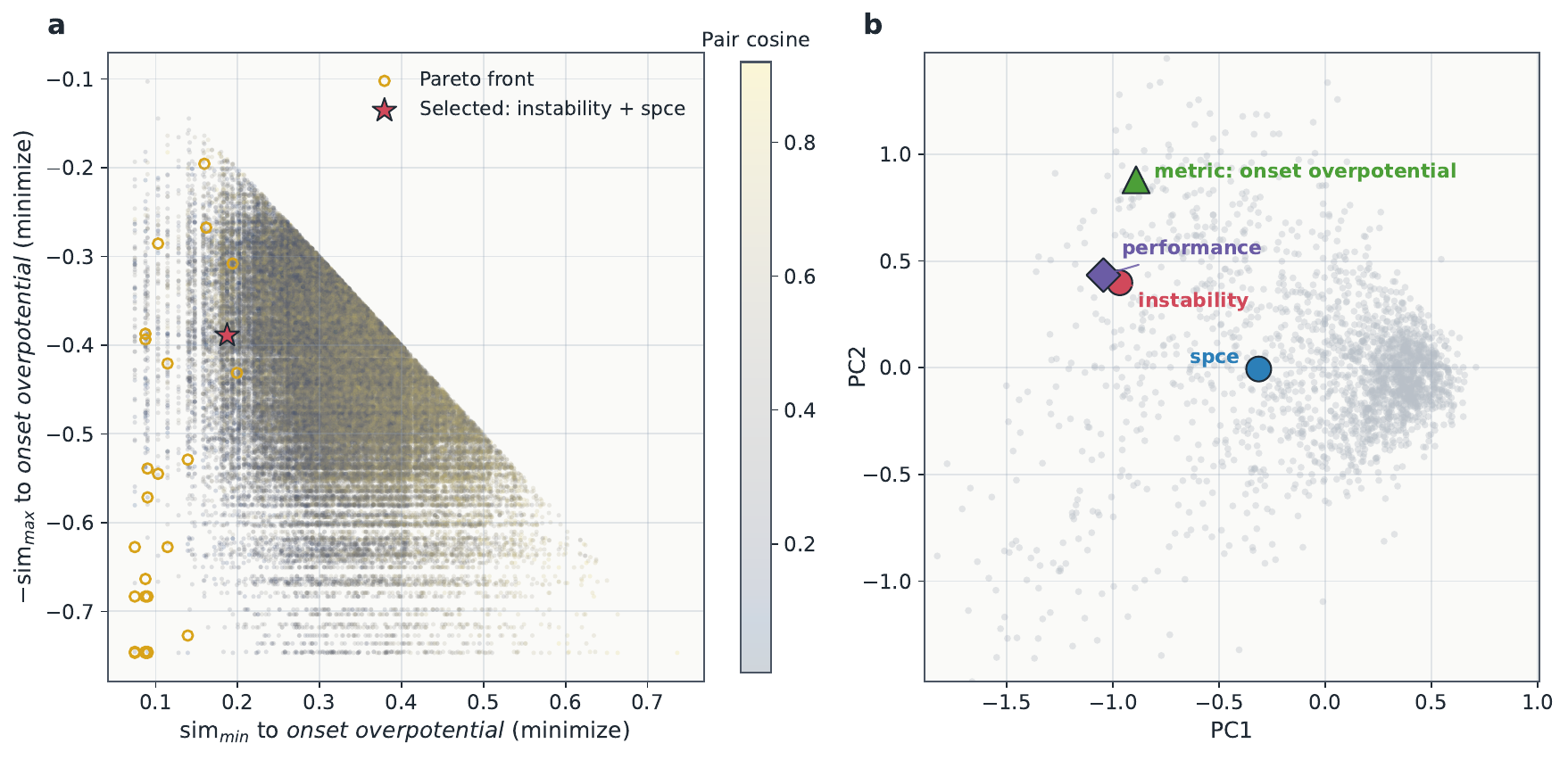}
\caption{Descriptor selection for the performance metric \textit{onset overpotential}. (a) Possible two-descriptor combinations in the three-objective selection space and the descriptor-level Pareto front. (b) PCA projection of the filtered embedding space, showing \textit{onset overpotential}, the word \textit{performance}, and the selected descriptors \textit{instability} and \textit{spce}.}
    \label{fig:si-concept-1}
\end{figure}

\begin{figure}[!htbp]
    \centering
    \includegraphics[width=\textwidth]{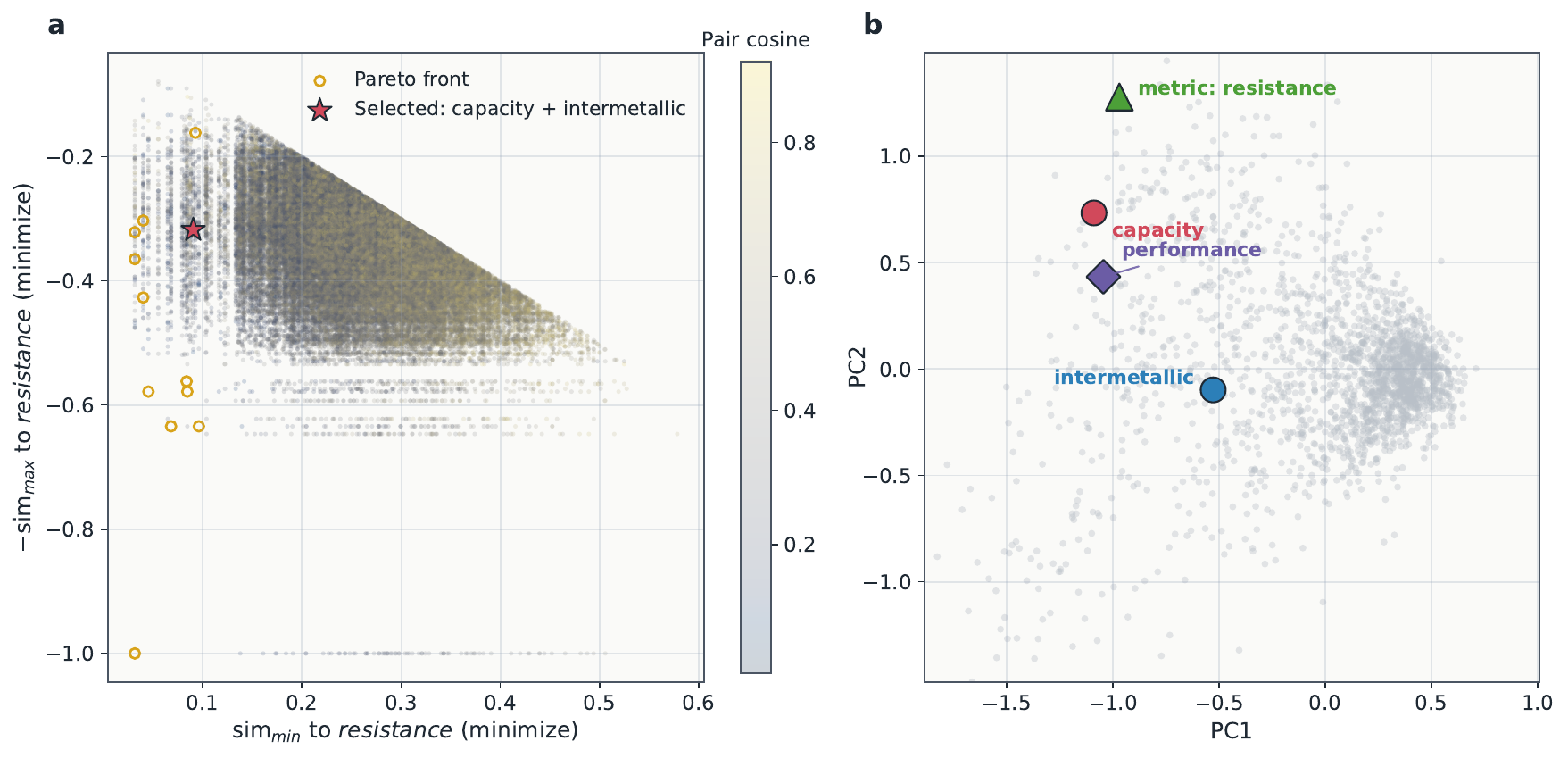}
\caption{Descriptor selection for the performance metric \textit{resistance}. (a) Possible two-descriptor combinations in the three-objective selection space and the descriptor-level Pareto front. (b) PCA projection of the filtered embedding space, showing \textit{resistance}, the word \textit{performance}, and the selected descriptors \textit{capacity} and \textit{intermetallic}.}
    \label{fig:si-concept-2}
\end{figure}

\begin{figure}[!htbp]
    \centering
    \includegraphics[width=\textwidth]{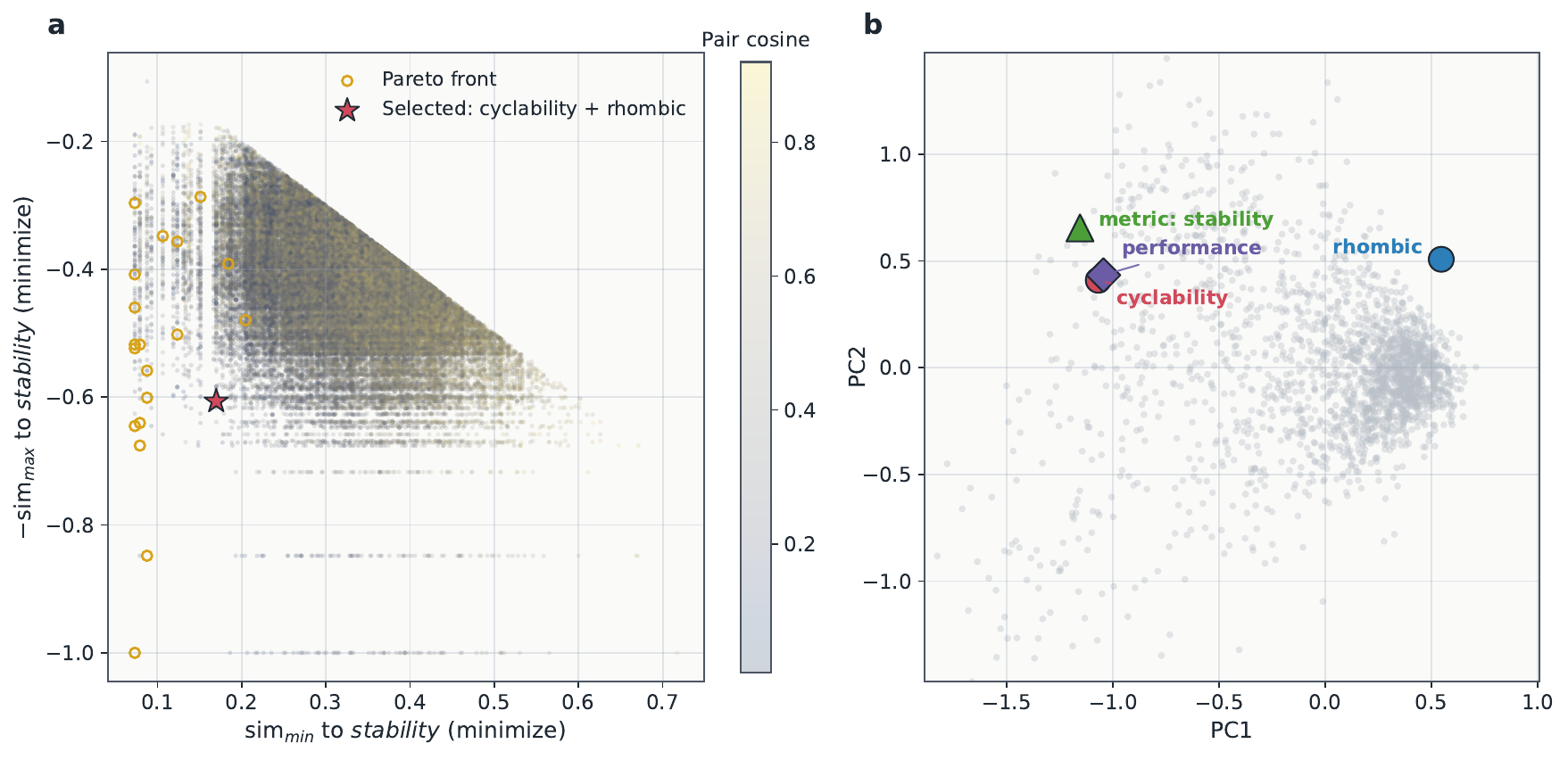}
\caption{Descriptor selection for the performance metric \textit{stability}. (a) Possible two-descriptor combinations in the three-objective selection space and the descriptor-level Pareto front. (b) PCA projection of the filtered embedding space, showing \textit{stability}, the word \textit{performance}, and the selected descriptors \textit{cyclability} and \textit{rhombic}.}
    \label{fig:si-concept-3}
\end{figure}

\end{document}